\documentclass{article} % For LaTeX2e
\usepackage{iclr2025_conference,times}

\usepackage{amsmath,amsfonts,bm}

\def\eqref#1{equation~\ref{#1}}
\def\1{\bm{1}}

\DeclareMathAlphabet{\mathsfit}{\encodingdefault}{\sfdefault}{m}{sl}
\SetMathAlphabet{\mathsfit}{bold}{\encodingdefault}{\sfdefault}{bx}{n}

\usepackage{amsmath}
\usepackage{hyperref}
\usepackage{url}
\usepackage{booktabs}
\usepackage{graphicx}
\usepackage{float}
\title{BanglaMamba: Exploring State Space Models for Bangla Fake News Detection}

\author{M. K. Khalidi Siam \\
Department of Computer Science and Engineering \\
BRAC University \\
Dhaka, Bangladesh \\
\texttt{\href{mailto:mk.khalidi.siam@g.bracu.ac.bd}{mk.khalidi.siam@g.bracu.ac.bd}}
}

\iclrfinalcopy % Uncomment for camera-ready version, but NOT for submission.
\begin{document}

\maketitle

\begin{abstract}
Fake news detection has become an important Natural Language Processing (NLP) task due to the rapid spread of misinformation through online news platforms and social media. While transformer-based models such as BanglaBERT achieve strong performance for Bangla text classification, their quadratic computational complexity makes them less suitable for long-document processing in resource-constrained environments. This paper investigates Mamba-based State Space Models (SSMs) as an efficient alternative for Bangla fake news detection. We propose BanglaMamba and compare it with pre-trained BanglaBERT and a similarly configured BERT model trained from scratch. Experimental results show that BanglaBERT achieves the highest Macro-F1 score (0.9260), while BanglaMamba (0.9029) achieves performance comparable to the from-scratch CustomBERT (0.9057) despite using a different architecture. Meanwhile, BanglaMamba achieves approximately $2.2\times$ higher inference throughput and 49\% lower inference peak GPU memory usage than the BERT-based models. Cross-dataset evaluation shows that BanglaBERT generalizes better to an external dataset, highlighting the importance of large-scale pretraining. These findings demonstrate that Mamba-based SSMs can provide a competitive and computationally efficient alternative to Transformer-based architectures for Bangla fake news detection, particularly in resource-constrained settings.
\end{abstract}

\section{Introduction}

Fake news has become a major challenge in the digital information era, where misleading information spreads rapidly through online news portals and social media. The widespread dissemination of false information can influence public opinion and undermine trust in credible news sources, making automatic fake news detection an important research problem in Natural Language Processing (NLP).

In recent years, transformer-based models such as BERT have significantly improved fake news detection by learning contextual representations that outperform traditional machine learning and conventional deep learning approaches~\citep{devlin-etal-2019-bert, Dell_Oglio_2026}. For Bangla NLP, BanglaBERT has emerged as a strong pretrained language model for text classification tasks~\citep{bhattacharjee-etal-2022-banglabert}.

Despite their effectiveness, transformer-based models have important limitations. The self-attention mechanism has quadratic time and memory complexity with respect to input length, resulting in increased computational cost and GPU memory consumption for longer sequences~\citep{NIPS2017_3f5ee243}.

Recent advances in State Space Models (SSMs), particularly the Mamba architecture, provide an alternative approach for efficient long-sequence modeling through linear scaling with sequence length while maintaining competitive sequence modeling performance~\citep{gu2024mambalineartimesequencemodeling}. However, the application of SSMs to Bangla fake news detection remains largely unexplored. Furthermore, existing studies have not systematically compared transformer-based and SSM-based models in terms of classification performance and computational efficiency.

To address these gaps, this paper investigates a Mamba-based State Space Model, referred to as BanglaMamba, for Bangla fake news detection and compares it with a pre-trained BanglaBERT and a similarly configured Transformer model trained from scratch (CustomBERT). The main contributions of this work are as follows:

\begin{itemize}

\item To the best of our knowledge, this is the first study to investigate a Mamba-based State Space Model (SSM) for Bangla fake news detection.

\item We provide a systematic comparison of BanglaMamba with BanglaBERT and a similarly configured CustomBERT in terms of classification performance and computational efficiency, including GPU memory usage, inference latency, and throughput, while using the from-scratch Transformer baseline to distinguish architectural effects from the benefits of language pretraining.

\item We evaluate the out-of-domain generalization of all three models on an external Bangla fake news dataset to examine their robustness under a shifted data distribution.

\end{itemize}

\section{Related Work}
\label{gen_inst}

Fake news detection has been extensively studied using traditional Machine Learning (ML), Deep Learning (DL), and, more recently, Transformer-based architectures across different languages. Early studies primarily relied on statistical machine learning models before gradually transitioning to contextual language models.

For Bangla fake news detection, \citet{10800650} constructed a dataset of fake and authentic Bangla news headlines and evaluated several machine learning algorithms alongside transformer-based models. Their results showed that BanglaBERT consistently outperformed all evaluated machine learning methods, demonstrating the effectiveness of contextual language representations.

Subsequent studies explored deep learning architectures. \citet{10.1007/978-3-031-73318-5_11} compared Bangla ELECTRA with a custom Deep CNN and reported that the CNN slightly outperformed the transformer model. Similarly, \citet{10055592} evaluated machine learning, deep learning, and transformer-based approaches for Bangla fake news detection and found that CNN achieved the highest classification performance, exceeding BanglaBERT by approximately 1\%. Nevertheless, BanglaBERT remained highly competitive, indicating the effectiveness of transformer-based models.

More recent studies have highlighted the superior generalization capability of transformers. \citet{Dell_Oglio_2026} compared machine learning, deep learning, transformer-based models, and large language models across ten English fake news datasets. While Logistic Regression and Support Vector Machine (SVM) performed competitively on smaller datasets, CNN and BiLSTM showed inconsistent performance and were more susceptible to overfitting. Fine-tuned transformer models achieved the best balance between classification accuracy and cross-domain generalization. Similar findings were reported by \citet{tabassum-etal-2024-punny} for Malayalam fake news detection. Additionally, \citet{Chowdhury_2025} showed that combining data augmentation with transformer fine-tuning significantly improved Bangla fake news detection performance.

Earlier Bangla fake news detection studies largely relied on handcrafted or shallow feature extraction techniques. Traditional machine learning methods commonly used Bag-of-Words (BoW), TF-IDF, and stemming, whereas deep learning models typically employed static Word2Vec embeddings~\citep{10800650, 10.1007/978-3-031-73318-5_11, 10055592}. Although computationally efficient, these approaches cannot effectively capture contextual meaning, long-range semantic dependencies, or word order~\citep{coenen2019visualizingmeasuringgeometrybert}. In contrast, transformer-based models automatically learn contextual embeddings using subword tokenization and self-attention~\citep{devlin-etal-2019-bert, clark2020electrapretrainingtextencoders, bhattacharjee-etal-2022-banglabert}. Their hierarchical representations capture lexical, syntactic, and semantic information across different layers~\citep{jawahar-etal-2019-bert, coenen2019visualizingmeasuringgeometrybert}, making transformers the dominant paradigm for modern Bangla NLP.

However, self-attention of transformer has quadratic time and memory complexity with respect to sequence length, making transformers computationally expensive for long-document processing~\citep{NIPS2017_3f5ee243}. State Space Models (SSMs) have emerged as an alternative for efficient long-sequence modeling, with Structured State Spaces (S4) demonstrating the ability to capture long-range dependencies while maintaining favorable computational efficiency~\citep{gu2022efficientlymodelinglongsequences}. Building on this line of work, Mamba introduces a selective state-space mechanism that enables input-dependent information propagation and achieves linear scaling in sequence length, while maintaining competitive performance on long sequences~\citep{gu2024mambalineartimesequencemodeling}. To the best of our knowledge, no published study has investigated SSMs for Bangla fake news detection. Moreover, existing studies have not systematically compared SSMs with BanglaBERT in terms of classification performance and computational efficiency.

\section{Methodology}
\label{headings}

\subsection{Problem Formulation}

This study addresses Bangla fake news detection as a binary text classification problem. Given a news article consisting of a headline and body text, the objective is to classify the article as Authentic (label = 1) or Fake (label = 0). Formally, the classifier learns a mapping function
\[
f \colon X \rightarrow \{0,1\},
\]
where \(X\) denotes the space of tokenized Bangla text sequences.

Two neural architectures and one reference baseline are evaluated:

\begin{itemize}
    \item \textbf{BanglaBERT}, a pre-trained Transformer encoder used as the primary baseline.

    \item \textbf{CustomBERT}, a Transformer model with a configuration comparable to BanglaBERT but trained entirely from scratch, providing a fair architecture-level comparison without large-scale pretraining.

    \item \textbf{BanglaMamba}, a Mamba-based State Space Model (SSM) trained entirely from scratch.
\end{itemize}

This study investigates the capability of the Mamba State Space Model architecture for Bangla fake news detection by comparing its classification performance with both a pre-trained Transformer baseline (BanglaBERT) and an architecturally comparable Transformer trained from scratch (CustomBERT). This comparison enables the analysis of the impact of model architecture while reducing the influence of large-scale language pretraining.

\subsection{Dataset and Preprocessing}
\subsubsection{Dataset}
All experiments are conducted using the BanFakeNews-2.0 dataset~\citep{shibu-etal-2025-scarcity}, a publicly available Bangla fake news corpus containing news articles annotated as Fake (0) or Real (1). Each sample consists of a news headline and its corresponding article body. To evaluate out-of-domain generalization, we additionally use the BanglaFakeNews2025 dataset~\citep{maliha2026banglafakenews2025} as an external evaluation dataset. Both datasets undergo the same preprocessing procedure to ensure consistency. After preprocessing, BanFakeNews-2.0 contains 58,001 samples, with 16.5\% labeled as Fake and 83.5\% as Real, while the external BanglaFakeNews2025 test set contains 3,979 samples, with 49.9\% Fake and 50.1\% Real.

\subsubsection{Preprocessing}
A preprocessing pipeline was applied to standardize the textual data before training. First, the headline and article body were concatenated using the \texttt{[SEP]} token to enable joint contextual learning. The preprocessing steps included: (i) removal of HTML tags and URLs using regular expressions, (ii) Unicode NFC normalization for consistent Bangla encoding, (iii) whitespace normalization by collapsing multiple spaces and trimming leading/trailing spaces, (iv) removal of articles with fewer than 20 or more than 2,000 words, (v) duplicate removal, and (vi) elimination of empty samples after preprocessing. Punctuation marks were preserved, as they may contain stylistic cues relevant to fake news detection. Approximately 5.5\% of the articles were removed as duplicates, while a further 0.3\% were removed because they contained fewer than 20 or more than 2,000 words.

The processed dataset was split into 80\% training, 10\% validation, and 10\% testing sets using stratified sampling to preserve the class distribution. Since the dataset was imbalanced, with approximately 83.5\% Real and 16.5\% Fake samples, inverse-frequency class weights were computed from the training set using Scikit-learn’s balanced class-weighting strategy and incorporated into the weighted cross-entropy loss during training.

\subsubsection{Context-Length Evaluation Sets}

Since all three models are evaluated with a maximum input length of 512 tokens, the held-out test set is divided into two subsets according to the BanglaBERT tokenizer to examine the effect of truncation:

\begin{itemize}
    \item \textbf{Short articles ($\leq 512$ tokens):} BanglaBERT, CustomBERT, and BanglaMamba process the complete article without truncation.

    \item \textbf{Long articles ($>512$ tokens):} BanglaBERT, CustomBERT, and BanglaMamba process only the first 512 tokens, with the remaining content truncated.
\end{itemize}

This evaluation examines whether the 512-token input constraint and the resulting truncation for longer articles substantially affect fake news detection performance across the three models.

\subsection{Tokenization}
All models use the BanglaBERT's SentencePiece tokenizer with a vocabulary size of 32{,}000 subword tokens. Using a common tokenizer ensures that differences in classification performance arise from the model architectures rather than differences in tokenization. Sequences shorter than the maximum sequence length are padded, while longer sequences are truncated according to each model's context limit.

\subsection{Model Architecture}

\subsubsection{BanglaBERT and CustomBERT}

\textbf{BanglaBERT} serves as the pre-trained Transformer baseline. It is a BERT-style Transformer encoder pre-trained on a large Bangla corpus. We use the smaller version of the BanglaBERT base model, referred to as BanglaBERT-small; throughout this paper, we simply refer to it as BanglaBERT~\citep{bhattacharjee-etal-2022-banglabert}. The exact pretrained checkpoint used in our experiments is publicly available on the Hugging Face Model Hub.\footnote{\url{https://huggingface.co/csebuetnlp/banglabert_small}} A linear classification layer is attached to the \texttt{[CLS]} token representation, and all parameters are fine-tuned end-to-end on the training set.

\textbf{CustomBERT} uses the same architecture and configuration as BanglaBERT but is trained entirely from scratch on the training data, without any large-scale language pretraining. This model is introduced to provide a more direct comparison with BanglaMamba by controlling for the effect of pretraining. Thus, the comparison between CustomBERT and BanglaMamba primarily reflects differences in the underlying Transformer and State Space Model architectures, while the comparison between BanglaBERT and CustomBERT illustrates the effect of pretraining under the same Transformer configuration.

\begin{table}[h]
\centering
\begin{tabular}{lll}
\hline
\textbf{Property} & \textbf{BanglaBERT} & \textbf{CustomBERT} \\
\hline
Architecture  & Transformer Encoder & Transformer Encoder \\
Parameters  & 13.7M & 13.7M \\
Hidden Dimension & 256 & 256 \\
Hidden Layers & 12 & 12 \\
Maximum Context Length & 512 Tokens & 512 Tokens \\
Pretraining & Yes & No \\
\hline
\end{tabular}
\caption{Configuration of BanglaBERT and CustomBERT.}
\label{tab:banglabert_config}
\end{table}

\subsubsection{BanglaMamba}
The proposed BanglaMamba model is built upon the Mamba State Space Model architecture and is trained entirely from scratch without using any pre-trained language representations. Since no publicly available pre-trained Mamba model is available within the parameter range of BanglaBERT (110M) and BanglaBERT-small (13.7M), we train BanglaMamba from scratch. To ensure a fair comparison with BanglaBERT, the model is configured with an almost similar parameter count to BanglaBERT.

\begin{table}[h]
\centering
\begin{tabular}{lll}
\hline
\textbf{Property} & \textbf{BanglaMamba} \\
\hline
Architecture  & Mamba \\
Parameters  & 12.61M \\
Hidden Dimension (d\_model) & 256 \\
Hidden Layer (n\_layer) & 10 \\
Maximum Context Length & 512 Tokens \\
\hline
\end{tabular}
\caption{Configuration of the BanglaMamba.}
\label{tab:banglamamba_config}
\end{table}

\begin{figure}[t]
    \centering
    \includegraphics[width=0.8\textwidth]{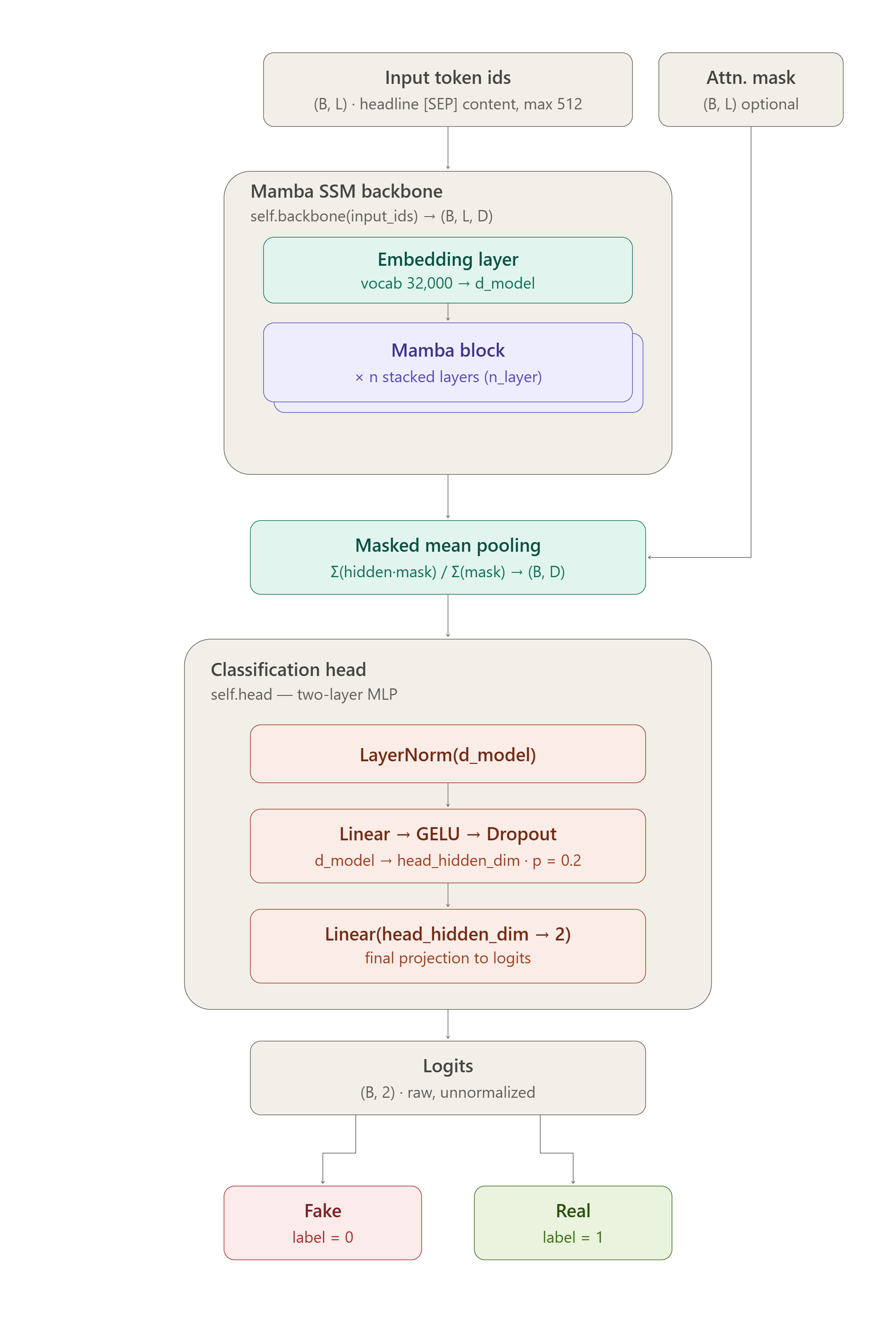}
    \caption{BanglaMamba model architecture.}
    \label{fig:banglamamba_arch}
\end{figure}

The architecture consists of three components, illustrated in 
Figure~\ref{fig:banglamamba_arch}:

\begin{enumerate}
    \item \textbf{Embedding Layer and Mamba Backbone}, which generate contextual token representations.
    
    \item \textbf{Masked Mean Pooling}, which computes a document-level representation by averaging hidden states corresponding only to valid tokens.
    
    \item \textbf{Classification Head}, consisting of:
    \[
    \text{LayerNorm} \rightarrow \text{Linear} \rightarrow \text{GELU} \rightarrow \text{Dropout}(0.2) \rightarrow \text{Linear}
    \]
    to perform binary classification.
\end{enumerate}

\subsection{Training Procedure}
All three models are trained using the weighted cross-entropy loss function and optimized with the AdamW optimizer. Since the three models use different architectures and initialization strategies, different learning rates are employed for each model. Training is conducted for 5 epochs using BF16 (bfloat16) mixed-precision computation to improve training efficiency while maintaining numerical stability. All experiments are performed on a single NVIDIA A100 GPU with 40 GB of memory.

At the end of each epoch, the models are evaluated on the validation set. The checkpoint achieving the highest Macro-F1 score on the validation set is selected as the final model and subsequently used for evaluation on the test set.

\subsection{Evaluation Methodology}

Due to the substantial class imbalance in BanFakeNews-2.0 (83.5\% authentic, 16.5\% fake), overall accuracy is not a reliable evaluation metric. Therefore, model performance is evaluated using multiple complementary metrics covering overall classification performance, class-wise discrimination, prediction reliability, and inference efficiency. In addition to predictive performance, inference efficiency is evaluated under a consistent configuration using a batch size of 64, \texttt{torch.no\_grad()} to disable gradient computation, and a maximum sequence length of 512 tokens for both models. All experiments are repeated using three random seeds (22, 33, and 42) to ensure reproducibility and reduce the influence of run-specific randomness. The results are averaged across the three runs to provide a more stable and reliable estimate of model performance, with the corresponding standard deviations also reported to quantify the variability across runs.

\subsubsection{Evaluation Metrics}

\textbf{Primary metric.} Macro-F1 is used as the primary evaluation metric because it gives equal weight to both classes, preventing the performance on the majority authentic class from masking performance on the minority fake class.

\textbf{Classification metrics.} F1-score are reported separately for the fake (label 0) and authentic (label 1) classes. AUC-ROC measures the model's threshold-independent discriminative ability, while PR-AUC provides a complementary evaluation that is particularly informative under class imbalance by emphasizing performance on the positive class.

\textbf{Reliability metrics.} Matthews Correlation Coefficient (MCC) is used as a robust measure of classification performance under class imbalance. Expected Calibration Error (ECE) is used to assess the alignment between predicted confidence and observed accuracy, computed using 10 equal-width bins over the confidence interval $[0,1]$:
\begin{equation}
\mathrm{ECE}=
\sum_{b=1}^{B}
\frac{|S_b|}{N}
\left|
\mathrm{acc}(S_b)-\mathrm{conf}(S_b)
\right|,
\end{equation}
where $B=10$, $S_b$ denotes the samples in bin $b$, and $\mathrm{acc}(S_b)$ and $\mathrm{conf}(S_b)$ represent the empirical accuracy and mean confidence, respectively.

\subsubsection{Efficiency Metrics}

\textbf{Peak GPU VRAM.} Measures the maximum GPU memory allocated during inference.

\textbf{P50 and P95 Latency.} P50 (median) latency represents typical inference latency per batch, while P95 latency captures tail inference latency per batch.

\textbf{Throughput (samples/s).} Measures the number of news articles processed per second during inference.

Complete details of the experimental setup, training, hyperparameter configurations, and detailed results are provided in the Appendix.

\section{Result Analysis}
\label{others}

\subsection{Classification Performance Comparison}

\begin{table}[ht]
\centering
\setlength{\tabcolsep}{12pt}
\renewcommand{\arraystretch}{1.2}
\begin{tabular}{lccc}
\toprule
\textbf{Metric} & \textbf{BanglaBERT} & \textbf{BanglaMamba} & \textbf{CustomBERT} \\
\midrule
Macro F1  & 0.9260 $\pm$ 0.0008 & 0.9029 $\pm$ 0.0034 & 0.9057 $\pm$ 0.0078 \\
AUC-ROC   & 0.9632 $\pm$ 0.0028 & 0.9420 $\pm$ 0.0103 & 0.9407 $\pm$ 0.0235 \\
MCC       & 0.8534 $\pm$ 0.0024 & 0.8086 $\pm$ 0.0085 & 0.8137 $\pm$ 0.0144 \\
PR-AUC    & 0.9896 $\pm$ 0.0013 & 0.9820 $\pm$ 0.0049 & 0.9810 $\pm$ 0.0097 \\
ECE       & 0.0507 $\pm$ 0.0371 & 0.0457 $\pm$ 0.0134 & 0.0537 $\pm$ 0.0176 \\
Fake F1   & 0.8753 $\pm$ 0.0011 & 0.8361 $\pm$ 0.0057 & 0.8408 $\pm$ 0.0136 \\
Real F1   & 0.9766 $\pm$ 0.0005 & 0.9697 $\pm$ 0.0017 & 0.9706 $\pm$ 0.0021 \\
\bottomrule
\end{tabular}
\caption{Classification performance comparison among BanglaBERT, BanglaMamba, and CustomBERT.}
\label{tab:classification_comparison}
\end{table}

Table~\ref{tab:classification_comparison} shows that BanglaBERT consistently achieves the highest classification performance across the evaluated metrics. In contrast, BanglaMamba and CustomBERT, both trained entirely from scratch with comparable model sizes, exhibit very similar performance. Their Macro-F1 scores are 0.9029 and 0.9057, respectively, while their Fake F1 scores are 0.8361 and 0.8408. The close results across the other metrics further indicate that neither model has a clear overall advantage when both are trained without large-scale language pretraining.

The performance gap between the two scratch-trained models and BanglaBERT suggests that pre-trained linguistic representations provide an important advantage for this classification task, particularly for the more challenging Fake class. However, the similar performance of BanglaMamba and CustomBERT indicates that, under comparable training conditions, the Mamba architecture can achieve classification performance comparable to a Transformer architecture of similar scale. This provides a more controlled comparison of the two architectures while separating the effect of large-scale pretraining from the effect of the underlying model architecture.

\subsection{Computational Efficiency Comparison}

\begin{table}[ht]
\centering
\renewcommand{\arraystretch}{1.2}
\resizebox{\textwidth}{!}{%
\begin{tabular}{lccc}
\toprule
\textbf{Metric} & \textbf{BanglaBERT} & \textbf{BanglaMamba} & \textbf{CustomBERT} \\
\midrule
P50 Latency (ms)         & 37.0767 $\pm$ 0.2701  & 14.8433 $\pm$ 2.2382   & 37.0333 $\pm$ 0.2845 \\
P95 Latency (ms)         & 39.03 $\pm$ 3.0967    & 17.4267 $\pm$ 3.1683   & 39.05 $\pm$ 3.12 \\
Throughput (sample/s)    & 1675.03 $\pm$ 24.8858 & 3623.2333 $\pm$ 125.4971 & 1680.13 $\pm$ 23.3548 \\
Inference Peak VRAM (MB) & 1011.9 $\pm$ 0        & 514.5 $\pm$ 0           & 1011.9 $\pm$ 0 \\
Training Peak VRAM (MB)  & 5141.6 $\pm$ 0        & 1145.4 $\pm$ 0          & 5141.6 $\pm$ 0 \\
\bottomrule
\end{tabular}%
}
\caption{Efficiency comparison among BanglaBERT, BanglaMamba, and CustomBERT.}
\label{tab:efficiency_comparison}
\end{table}

Table~\ref{tab:efficiency_comparison} compares the efficiency of BanglaBERT, BanglaMamba, and CustomBERT across inference latency, throughput, and memory consumption. BanglaMamba demonstrates substantial improvements in inference speed and processing capacity, achieving approximately $2.5\times$ lower P50 latency and $2.2\times$ lower P95 latency than BanglaBERT. At the same time, it achieves approximately $2.2\times$ higher throughput, indicating that it can process substantially more samples per second under the same evaluation configuration. Although BanglaMamba exhibits higher variability in throughput across runs, its average throughput remains considerably higher than that of BanglaBERT.

Since CustomBERT uses exactly the same architecture and configuration as BanglaBERT, its efficiency metrics are effectively identical. Overall, BanglaMamba demonstrates a clear efficiency advantage, requiring approximately $49\%$ less peak VRAM during inference and nearly $78\%$ less during training. The lower latency, higher throughput, and reduced memory consumption are consistent with the computational advantages of its linear-time sequence processing, highlighting its potential for resource-constrained deployment.

\subsection{Input Length Robustness Analysis}

\begin{table}[ht]
\centering
\renewcommand{\arraystretch}{1.2}
\resizebox{\textwidth}{!}{%
\begin{tabular}{lcccccc}
\toprule
\textbf{Metric} &
\multicolumn{2}{c}{\textbf{BanglaBERT}} &
\multicolumn{2}{c}{\textbf{BanglaMamba}} &
\multicolumn{2}{c}{\textbf{CustomBERT}} \\
\cmidrule(lr){2-3}
\cmidrule(lr){4-5}
\cmidrule(lr){6-7}
& \textbf{Short} & \textbf{Long}
& \textbf{Short} & \textbf{Long}
& \textbf{Short} & \textbf{Long} \\
\midrule

Macro F1
& 0.9309 $\pm$ 0.0047
& 0.9054 $\pm$ 0.0175
& 0.9030 $\pm$ 0.0012
& 0.9012 $\pm$ 0.0217
& 0.9049 $\pm$ 0.0063
& 0.9080 $\pm$ 0.0149 \\

AUC-ROC
& 0.9662 $\pm$ 0.0062
& 0.9466 $\pm$ 0.0193
& 0.9435 $\pm$ 0.0088
& 0.9351 $\pm$ 0.0201
& 0.9424 $\pm$ 0.0213
& 0.9336 $\pm$ 0.0363 \\

MCC
& 0.8633 $\pm$ 0.0099
& 0.8127 $\pm$ 0.0342
& 0.8092 $\pm$ 0.0046
& 0.8042 $\pm$ 0.0427
& 0.8120 $\pm$ 0.0122
& 0.8196 $\pm$ 0.0253 \\

Fake F1
& 0.8831 $\pm$ 0.0082
& 0.8427 $\pm$ 0.0323
& 0.8357 $\pm$ 0.0024
& 0.8351 $\pm$ 0.0391
& 0.8390 $\pm$ 0.0106
& 0.8462 $\pm$ 0.0284 \\

\bottomrule
\end{tabular}%
}
\caption{Performance comparison of BanglaBERT, BanglaMamba, and CustomBERT on short and long articles.}
\label{tab:input_robustness}
\end{table}

Table~\ref{tab:input_robustness} compares model performance on short and long articles when all models are restricted to a maximum input length of 512 tokens. Overall, the performance differences are relatively small, indicating that truncation of longer articles does not substantially degrade classification performance. BanglaBERT shows a somewhat larger decrease across the evaluated metrics, while BanglaMamba and CustomBERT exhibit smaller changes, with CustomBERT even showing a slight improvement in Macro-F1 and Fake F1 on long articles. These results suggest that the 512-token input constraint does not significantly affect the overall classification behavior of the models on longer articles.

\subsection{Cross-Dataset Generalization}

\begin{table}[ht]
\centering
\setlength{\tabcolsep}{12pt}
\renewcommand{\arraystretch}{1.2}
\begin{tabular}{lccc}
\toprule
\textbf{Metric} & \textbf{BanglaBERT} & \textbf{BanglaMamba} & \textbf{CustomBERT} \\
\midrule
Macro F1  & 0.7914 $\pm$ 0.0348 & 0.5753 $\pm$ 0.0703 & 0.7227 $\pm$ 0.0892 \\
AUC-ROC   & 0.8693 $\pm$ 0.0185 & 0.7598 $\pm$ 0.0164 & 0.7777 $\pm$ 0.1218 \\
MCC       & 0.5913 $\pm$ 0.0708 & 0.2590 $\pm$ 0.0767 & 0.4746 $\pm$ 0.1696 \\
PR-AUC    & 0.8622 $\pm$ 0.0076 & 0.7845 $\pm$ 0.0172 & 0.7197 $\pm$ 0.1502 \\
ECE       & 0.0929 $\pm$ 0.0423 & 0.2564 $\pm$ 0.0472 & 0.1448 $\pm$ 0.0605 \\
Fake F1   & 0.7763 $\pm$ 0.0365 & 0.4564 $\pm$ 0.1253 & 0.6824 $\pm$ 0.1081 \\
Real F1   & 0.8064 $\pm$ 0.0337 & 0.6942 $\pm$ 0.0154 & 0.7630 $\pm$ 0.0706 \\
\bottomrule
\end{tabular}
\caption{Classification performance among BanglaBERT, BanglaMamba, and CustomBERT on the cross-dataset.}
\label{tab:bert_mam_performance}
\end{table}

To assess out-of-domain generalization, all three models were evaluated on an external Bangla fake news dataset that was not used during training. Out-of-distribution (OOD) generalization remains a known challenge in NLP when the evaluation distribution differs from the training distribution~\citep{yang-etal-2023-distribution}. As shown in Table~\ref{tab:bert_mam_performance}, BanglaBERT achieves the strongest overall OOD performance, followed by CustomBERT and BanglaMamba. In particular, BanglaBERT achieves substantially higher Macro-F1, MCC, and Fake F1 than BanglaMamba, while CustomBERT also outperforms BanglaMamba across these classification metrics. However, CustomBERT exhibits noticeably larger standard deviations across several metrics, indicating greater variation in performance across runs. In contrast, BanglaMamba shows lower variability for several metrics, although its overall classification performance remains substantially weaker.

The weaker OOD performance of the two models trained from scratch, particularly BanglaMamba, suggests that the absence of large-scale language pretraining may limit generalization beyond the downstream training distribution. This interpretation is consistent with prior work showing that effective language-model training depends strongly on an appropriate balance between model capacity and the amount of training data~\citep{hoffmann2022trainingcomputeoptimallargelanguage}. However, the observed performance gap cannot be attributed solely to the absence of pretraining, as architectural and optimization differences may also contribute. These results motivate large-scale pretraining of BanglaMamba on diverse Bangla text as an important direction for future work.

\subsection{Ablation Study}
\begin{table}[ht]
\centering
\setlength{\tabcolsep}{12pt}
\renewcommand{\arraystretch}{1.2}
\begin{tabular}{lcc}
\toprule
\textbf{Metric} & \textbf{BanglaBERT (Mean Pooling)} \\
\midrule
Macro F1 & 0.9144 $\pm$ 0.0081 \\
AUC-ROC  & 0.9565 $\pm$ 0.0077 \\
MCC      & 0.8300 $\pm$ 0.0172 \\
PR-AUC   & 0.9886 $\pm$ 0.0019 \\
ECE      & 0.0866 $\pm$ 0.0182 \\
Fake F1  & 0.8565 $\pm$ 0.0124 \\
Real F1  & 0.9723 $\pm$ 0.0038 \\
\bottomrule
\end{tabular}
\caption{Classification performance of BanglaBERT using mean pooling instead of the \texttt{[CLS]} representation.}
\label{tab:banglabert_mean_pooling}
\end{table}

Our main experiments use masked mean pooling for BanglaMamba and the \texttt{[CLS]} representation for BanglaBERT. To examine whether the pooling strategy substantially affects the BanglaBERT results, we additionally evaluate BanglaBERT using masked mean pooling, with the results shown in Table~\ref{tab:banglabert_mean_pooling}. The performance remains broadly comparable to the original \texttt{[CLS]}-based approach, although \texttt{[CLS]} pooling achieves slightly better overall results. This is consistent with the fact that BanglaBERT was pre-trained with the \texttt{[CLS]} representation as the sequence-level representation.

We do not perform a corresponding \texttt{[CLS]} ablation for BanglaMamba because the two architectures use fundamentally different sequence-processing mechanisms. The \texttt{[CLS]} representation is specifically designed for BERT-style bidirectional self-attention, whereas BanglaMamba processes sequences causally. Masked mean pooling therefore provides a more architecture-independent sequence representation for our comparison.

\section{Limitations and Future Work}

Although both BanglaBERT and BanglaMamba achieve strong in-domain performance, the class imbalance in the BanFakeNews-2.0 dataset remains a limitation. Both models exhibit lower performance on the minority (Fake) class than on the majority (Real) class, indicating that identifying fake news remains the more challenging aspect of the task. A key limitation of this study is that BanglaMamba is trained entirely from scratch using only the downstream fake news dataset. Due to computational resource constraints, we were unable to perform large-scale language pretraining. The cross-dataset evaluation further shows that BanglaMamba performs substantially worse than BanglaBERT on an external dataset, suggesting that the absence of large-scale pretraining may limit its out-of-domain generalization. However, this performance gap cannot be attributed solely to the lack of pretraining, as architectural and optimization differences may also contribute.

Future work should investigate large-scale pretraining of BanglaMamba on diverse Bangla text and evaluate the resulting model across multiple Bangla fake news datasets. Such experiments would provide a stronger assessment of the generalization capability and potential of pretrained Mamba-based models for Bangla fake news detection.

\section{Conclusion}

This study investigated Mamba-based State Space Models (SSMs) for Bangla fake news detection by comparing BanglaMamba with BanglaBERT and a similarly configured CustomBERT trained from scratch. Although BanglaBERT achieves the highest in-domain classification performance, BanglaMamba achieves competitive results despite being trained entirely from scratch, while providing substantially lower inference latency, higher throughput, and lower GPU memory usage. The similar in-domain performance of BanglaMamba and CustomBERT also suggests that training from scratch can produce competitive classification performance despite the absence of large-scale language pretraining.

Cross-dataset evaluation reveals weaker out-of-domain generalization for both models trained from scratch, with BanglaMamba showing the largest performance degradation. These results highlight the potential importance of large-scale language pretraining for improving generalization, while indicating that the observed gap cannot be attributed solely to model architecture. Overall, the findings demonstrate that Mamba-based SSMs are a promising and computationally efficient alternative to Transformer-based models for Bangla fake news detection. Future work should investigate large-scale pretraining of BanglaMamba and evaluate it across multiple Bangla fake news datasets.

\bibliography{iclr2025_conference}

@INPROCEEDINGS{10800650,
  author={Khatun, Mst. Sadia and Khan, Ishmam},
  booktitle={2024 IEEE International Conference on Power, Electrical, Electronics and Industrial Applications (PEEIACON)}, 
  title={Bangla Counterfeit News Identification: Using the Power of BERT}, 
  year={2024},
  volume={},
  number={},
  pages={518-522},
  doi={10.1109/PEEIACON63629.2024.10800650}}

@InProceedings{10.1007/978-3-031-73318-5_11,
author="Habiba, Sultana Umme
and Johora, Fatema Tuj
and Mahmud, Tanjim
and Tasnim, Tanpia
and Tasnim, Farzana
and Kabir, Sumaiya
and Hossain, Mohammad Shahadat
and Andersson, Karl",
editor="Vasant, Pandian
and Panchenko, Vladimir
and Munapo, Elias
and Weber, Gerhard-Wilhelm
and Thomas, J. Joshua
and Intan, Rolly
and Shamsul Arefin, Mohammad",
title="Enhancing Low-Resource Bangla Fake News Detection through Deep Convolutional Neural Networks",
booktitle="Intelligent Computing and Optimization",
year="2024",
publisher="Springer Nature Switzerland",
address="Cham",
pages="104--114",
isbn="978-3-031-73318-5"
}

@INPROCEEDINGS{10055592,
  author={Rasel, Risul Islam and Zihad, Anower Hossen and Sultana, Nasrin and Hoque, Mohammed Moshiul},
  booktitle={2022 25th International Conference on Computer and Information Technology (ICCIT)}, 
  title={Bangla Fake News Detection using Machine Learning, Deep Learning and Transformer Models}, 
  year={2022},
  volume={},
  number={},
  pages={959-964},
  doi={10.1109/ICCIT57492.2022.10055592}}

@article{Dell_Oglio_2026,
   title={An experimental comparison of the most popular approaches to fake news detection},
   volume={745},
   ISSN={0020-0255},
   url={http://dx.doi.org/10.1016/j.ins.2026.123407},
   DOI={10.1016/j.ins.2026.123407},
   journal={Information Sciences},
   publisher={Elsevier BV},
   author={Dell’Oglio, Pietro and Bondielli, Alessandro and Marcelloni, Francesco and Passaro, Lucia C.},
   year={2026},
   month=July, pages={123407} 
}

@article{Chowdhury_2025,
   title={Tackling Fake News in Bengali: Unraveling the Impact of Summarization vs. Augmentation on Pre-trained Language Models},
   volume={6},
   ISSN={2661-8907},
   url={http://dx.doi.org/10.1007/s42979-025-04398-z},
   DOI={10.1007/s42979-025-04398-z},
   number={8},
   journal={SN Computer Science},
   publisher={Springer Science and Business Media LLC},
   author={Chowdhury, Arman Sakif and Shahariar, G. M. and Aziz, Ahammed Tarik and Alam, Syed Mohibul and Sheikh, Md. Azad and Belal, Tanveer Ahmed},
   year={2025},
   month=Oct }

@inproceedings{tabassum-etal-2024-punny,
    title = "{P}unny{\_}{P}unctuators@{D}ravidian{L}ang{T}ech-{EACL}2024: Transformer-based Approach for Detection and Classification of Fake News in {M}alayalam Social Media Text",
    author = "Tabassum, Nafisa  and
      Aodhora, Sumaiya  and
      Akter, Rowshon  and
      Hossain, Jawad  and
      Ahsan, Shawly  and
      Hoque, Mohammed Moshiul",
    editor = "Chakravarthi, Bharathi Raja  and
      Priyadharshini, Ruba  and
      Madasamy, Anand Kumar  and
      Thavareesan, Sajeetha  and
      Sherly, Elizabeth  and
      Nadarajan, Rajeswari  and
      Ravikiran, Manikandan",
    booktitle = "Proceedings of the Fourth Workshop on Speech, Vision, and Language Technologies for Dravidian Languages",
    month = mar,
    year = "2024",
    address = "St. Julian's, Malta",
    publisher = "Association for Computational Linguistics",
    url = "https://aclanthology.org/2024.dravidianlangtech-1.30/",
    doi = "10.18653/v1/2024.dravidianlangtech-1.30",
    pages = "180--186"
}

@inproceedings{devlin-etal-2019-bert,
    title = "{BERT}: Pre-training of Deep Bidirectional Transformers for Language Understanding",
    author = "Devlin, Jacob  and
      Chang, Ming-Wei  and
      Lee, Kenton  and
      Toutanova, Kristina",
    editor = "Burstein, Jill  and
      Doran, Christy  and
      Solorio, Thamar",
    booktitle = "Proceedings of the 2019 Conference of the North {A}merican Chapter of the Association for Computational Linguistics: Human Language Technologies, Volume 1 (Long and Short Papers)",
    month = jun,
    year = "2019",
    address = "Minneapolis, Minnesota",
    publisher = "Association for Computational Linguistics",
    url = "https://aclanthology.org/N19-1423/",
    doi = "10.18653/v1/N19-1423",
    pages = "4171--4186"
}

@inproceedings{bhattacharjee-etal-2022-banglabert,
    title = "{B}angla{BERT}: Language Model Pretraining and Benchmarks for Low-Resource Language Understanding Evaluation in {B}angla",
    author = "Bhattacharjee, Abhik  and
      Hasan, Tahmid  and
      Ahmad, Wasi  and
      Mubasshir, Kazi Samin  and
      Islam, Md Saiful  and
      Iqbal, Anindya  and
      Rahman, M. Sohel  and
      Shahriyar, Rifat",
    editor = "Carpuat, Marine  and
      de Marneffe, Marie-Catherine  and
      Meza Ruiz, Ivan Vladimir",
    booktitle = "Findings of the Association for Computational Linguistics: NAACL 2022",
    month = jul,
    year = "2022",
    address = "Seattle, United States",
    publisher = "Association for Computational Linguistics",
    url = "https://aclanthology.org/2022.findings-naacl.98/",
    doi = "10.18653/v1/2022.findings-naacl.98",
    pages = "1318--1327"
}

@misc{clark2020electrapretrainingtextencoders,
      title={ELECTRA: Pre-training Text Encoders as Discriminators Rather Than Generators}, 
      author={Kevin Clark and Minh-Thang Luong and Quoc V. Le and Christopher D. Manning},
      year={2020},
      eprint={2003.10555},
      archivePrefix={arXiv},
      primaryClass={cs.CL},
      url={https://arxiv.org/abs/2003.10555}, 
}

@inproceedings{jawahar-etal-2019-bert,
    title = "What Does {BERT} Learn about the Structure of Language?",
    author = "Jawahar, Ganesh  and
      Sagot, Beno{\^i}t  and
      Seddah, Djam{\'e}",
    editor = "Korhonen, Anna  and
      Traum, David  and
      M{\`a}rquez, Llu{\'i}s",
    booktitle = "Proceedings of the 57th Annual Meeting of the Association for Computational Linguistics",
    month = jul,
    year = "2019",
    address = "Florence, Italy",
    publisher = "Association for Computational Linguistics",
    url = "https://aclanthology.org/P19-1356/",
    doi = "10.18653/v1/P19-1356",
    pages = "3651--3657"
}

@misc{coenen2019visualizingmeasuringgeometrybert,
      title={Visualizing and Measuring the Geometry of BERT}, 
      author={Andy Coenen and Emily Reif and Ann Yuan and Been Kim and Adam Pearce and Fernanda Viégas and Martin Wattenberg},
      year={2019},
      eprint={1906.02715},
      archivePrefix={arXiv},
      primaryClass={cs.LG},
      url={https://arxiv.org/abs/1906.02715}, 
}

@misc{gu2024mambalineartimesequencemodeling,
      title={Mamba: Linear-Time Sequence Modeling with Selective State Spaces}, 
      author={Albert Gu and Tri Dao},
      year={2024},
      eprint={2312.00752},
      archivePrefix={arXiv},
      primaryClass={cs.LG},
      url={https://arxiv.org/abs/2312.00752}, 
}

@inproceedings{shibu-etal-2025-scarcity,
    title = "From Scarcity to Capability: Empowering Fake News Detection in Low-Resource Languages with {LLM}s",
    author = "Shibu, Hrithik Majumdar  and
      Datta, Shrestha  and
      Miah, Md. Sumon  and
      Sami, Nasrullah  and
      Chowdhury, Mahruba Sharmin  and
      Islam, Md Saiful",
    editor = "Weerasinghe, Ruvan  and
      Anuradha, Isuri  and
      Sumanathilaka, Deshan",
    booktitle = "Proceedings of the First Workshop on Natural Language Processing for Indo-Aryan and Dravidian Languages",
    month = jan,
    year = "2025",
    address = "Abu Dhabi",
    publisher = "Association for Computational Linguistics",
    url = "https://aclanthology.org/2025.indonlp-1.12/",
    pages = "100--107"
}

@misc{hoffmann2022trainingcomputeoptimallargelanguage,
      title={Training Compute-Optimal Large Language Models}, 
      author={Jordan Hoffmann and Sebastian Borgeaud and Arthur Mensch and Elena Buchatskaya and Trevor Cai and Eliza Rutherford and Diego de Las Casas and Lisa Anne Hendricks and Johannes Welbl and Aidan Clark and Tom Hennigan and Eric Noland and Katie Millican and George van den Driessche and Bogdan Damoc and Aurelia Guy and Simon Osindero and Karen Simonyan and Erich Elsen and Jack W. Rae and Oriol Vinyals and Laurent Sifre},
      year={2022},
      eprint={2203.15556},
      archivePrefix={arXiv},
      primaryClass={cs.CL},
      url={https://arxiv.org/abs/2203.15556}, 
}

@inproceedings{yang-etal-2023-distribution,
    title = "Out-of-Distribution Generalization in Natural Language Processing: Past, Present, and Future",
    author = "Yang, Linyi  and
      Song, Yaoxian  and
      Ren, Xuan  and
      Lyu, Chenyang  and
      Wang, Yidong  and
      Zhuo, Jingming  and
      Liu, Lingqiao  and
      Wang, Jindong  and
      Foster, Jennifer  and
      Zhang, Yue",
    editor = "Bouamor, Houda  and
      Pino, Juan  and
      Bali, Kalika",
    booktitle = "Proceedings of the 2023 Conference on Empirical Methods in Natural Language Processing",
    month = dec,
    year = "2023",
    address = "Singapore",
    publisher = "Association for Computational Linguistics",
    url = "https://aclanthology.org/2023.emnlp-main.276/",
    doi = "10.18653/v1/2023.emnlp-main.276",
    pages = "4533--4559"
}

@misc{maliha2026banglafakenews2025,
  author       = {Maliha, Samia and Hossain, Md. Shahrukh and Anwar, Md. Musfique and Seema, Sharmeen Jahan},
  title        = {{BanglaFakeNews2025}: A High-Quality Benchmark Dataset for Bangla Fake News Detection},
  year         = {2026},
  publisher    = {Mendeley Data},
  version      = {V3},
  doi          = {10.17632/c6hf7g5f3t.3}
}

@inproceedings{NIPS2017_3f5ee243,
 author = {Vaswani, Ashish and Shazeer, Noam and Parmar, Niki and Uszkoreit, Jakob and Jones, Llion and Gomez, Aidan N and Kaiser, \L ukasz and Polosukhin, Illia},
 booktitle = {Advances in Neural Information Processing Systems},
 editor = {I. Guyon and U. Von Luxburg and S. Bengio and H. Wallach and R. Fergus and S. Vishwanathan and R. Garnett},
 pages = {},
 publisher = {Curran Associates, Inc.},
 title = {Attention is All you Need},
 url = {https://proceedings.neurips.cc/paper_files/paper/2017/file/3f5ee243547dee91fbd053c1c4a845aa-Paper.pdf},
 volume = {30},
 year = {2017}
}

@misc{gu2022efficientlymodelinglongsequences,
      title={Efficiently Modeling Long Sequences with Structured State Spaces}, 
      author={Albert Gu and Karan Goel and Christopher Ré},
      year={2022},
      eprint={2111.00396},
      archivePrefix={arXiv},
      primaryClass={cs.LG},
      url={https://arxiv.org/abs/2111.00396}, 
}
\bibliographystyle{iclr2025_conference}

\section*{AI Usage Statement}
We used AI-assisted tools to improve the quality of our writing, including grammatical correction and language polishing. We also used AI tools to assist with code development. All experimental data were generated through our own experiments and pipelines. No results, claims, or empirical findings were generated by AI tools. All interpretations, analyses, and conclusions were produced by the authors, with AI tools used solely to polish the presentation of the analysis.

\appendix
\section{Experimental Setup}

\begin{table}[H]
\centering
\begin{tabular}{ll}
\hline
\textbf{Component} & \textbf{Specification} \\
\hline
Programming Language & Python 3.11 \\
Deep Learning Framework & PyTorch 2.4.0 + CUDA 12.4 \\
Transformer Library & Transformers 4.40.0 \\
State Space Model Library & Mamba-SSM 2.2.2 \\
GPU & NVIDIA A100 (40 GB) \\
\hline
\end{tabular}
\caption{Experimental setup.}
\label{tab:experimental_setup}
\end{table}

\section{BanglaMamba Model Configuration}

\begin{table}[H]
\centering
\begin{tabular}{lc}
\toprule
\textbf{Configuration} & \textbf{Value} \\
\midrule
$d_{\mathrm{model}}$ & 256 \\
$head\_hidden\_dim$ & 128 \\
$n_{\mathrm{layer}}$ & 10 \\
Dropout & 0.2 \\
Parameters & 12.61M \\
\bottomrule
\end{tabular}
\caption{Configurations of the BanglaMamba model.}
\label{tab:mamba_configurations}
\end{table}

The Mamba-specific parameters were kept unchanged from the default configuration of the official \texttt{mamba-ssm} package. Specifically, the state dimension ($d_{\mathrm{state}}$) was set to 16, the convolutional kernel size ($d_{\mathrm{conv}}$) was set to 4, and the expansion factor (\texttt{expand}) was set to 2. The $\Delta$ rank (\texttt{dt\_rank}) was set to \texttt{auto}, calculated as $\lceil d_{\mathrm{model}}/16 \rceil$. In addition, \texttt{use\_bias} was set to \texttt{False} and \texttt{conv\_bias} was set to \texttt{True}, following the default settings of the official \texttt{mamba-ssm} implementation.

\section{Training Configuration}
\begin{table}[H]
\centering
\begin{tabular}{lccc}
\hline
\textbf{Hyperparameter} & \textbf{BanglaBERT} & \textbf{BanglaMamba} & \textbf{CustomBERT} \\
\hline
Vocabulary Size & 32,000 & 32,000 & 32,000 \\
Maximum Sequence Length & 512 & 512 & 512 \\
Number of Labels & 2 & 2 & 2 \\
Epochs & 5 & 5 & 5 \\
Batch Size & 32 & 32 & 32 \\
Gradient Accumulation & 2 & 2 & 2 \\
Effective Batch Size & 64 & 64 & 64 \\
Learning Rate & $2\times10^{-5}$ & $1\times10^{-3}$ & $2\times10^{-4}$ \\
Weight Decay & 0.01 & 0.01 & 0.01 \\
Warm-up Ratio & 0.10 & 0.10 & 0.10 \\
Precision & BF16 & BF16 & BF16 \\
\hline
\end{tabular}
\caption{Training configurations for BanglaBERT, BanglaMamba, and CustomBERT.}
\label{tab:training_config}
\end{table}

\subsection{Training Dynamics}
\label{app:training_dynamics}

\begin{figure}[ht]
    \centering
    \includegraphics[width=\textwidth]{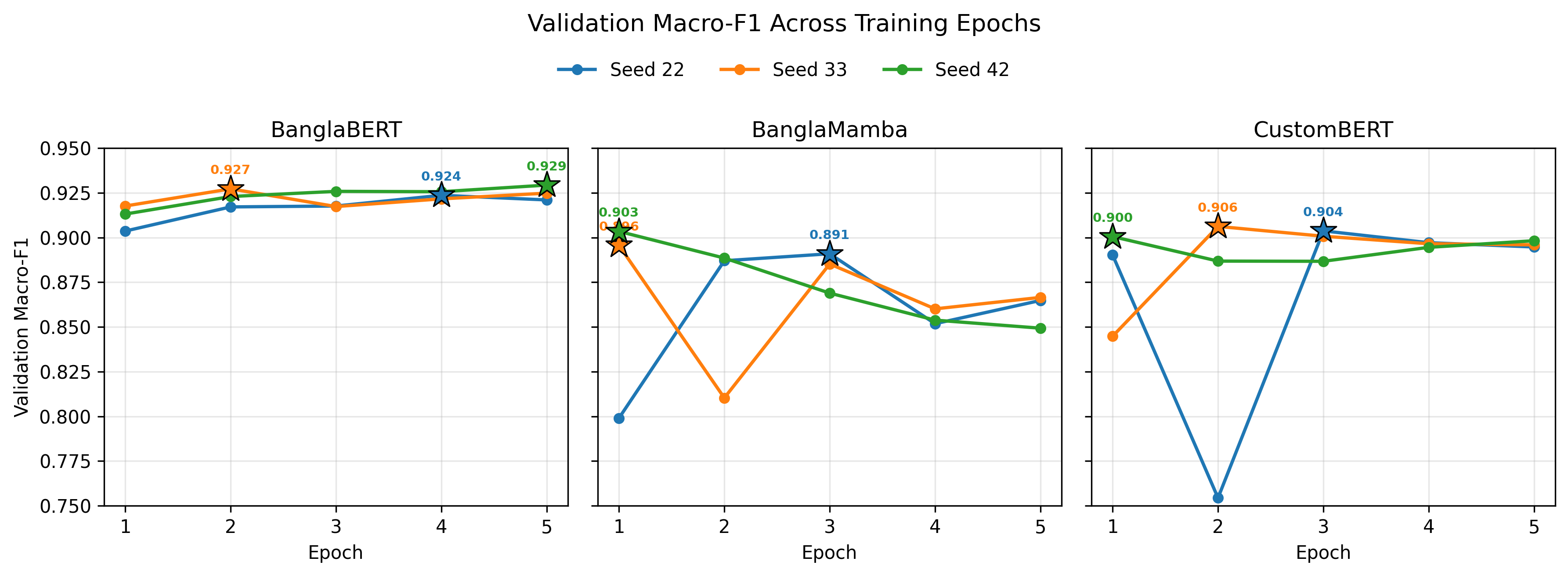}
    \caption{Validation Macro-F1 across training epochs for BanglaBERT, BanglaMamba, and CustomBERT over three random seeds.}
    \label{fig:training_dynamics}
\end{figure}

Figure~\ref{fig:training_dynamics} presents the validation Macro-F1 across the five training epochs for BanglaBERT, BanglaMamba, and CustomBERT over three random seeds. The models were trained using the learning rates reported in Table~\ref{tab:training_config}. We also evaluated alternative learning rates for each model, but none produced better validation performance than the selected configurations. For the pretrained BanglaBERT, the lower learning rate of $2\times10^{-5}$ provided the most stable and effective fine-tuning, whereas larger learning rates did not improve performance. For BanglaMamba, trained entirely from scratch, $10^{-3}$ provided the best performance. In contrast, applying $10^{-3}$ to CustomBERT resulted in ineffective optimization, with the model collapsing to the majority class and obtaining a validation Macro-F1 of only 0.455 across all five epochs. Reducing the learning rate to $2\times10^{-4}$ substantially improved CustomBERT's performance.

Under the selected learning rates, all three models generally reached their best validation performance within the first few epochs, with limited or inconsistent improvement afterward. BanglaBERT improved more gradually, while BanglaMamba and CustomBERT generally reached their best performance earlier. These training dynamics, together with the lack of meaningful improvement in later epochs, support the use of five training epochs for the reported experiments.

\section{Additional Experimental Results}
\label{app:additional_results}

\subsection{BanglaBERT Results}
\begin{table}[H]
\centering
\setlength{\tabcolsep}{10pt}
\renewcommand{\arraystretch}{1.15}
\begin{tabular}{lccc}
\toprule
\textbf{Metric} & \textbf{Seed 22} & \textbf{Seed 33} & \textbf{Seed 42} \\
\midrule
Macro F1 & 0.9255 & 0.9255 & 0.9269 \\
AUC-ROC  & 0.9607 & 0.9626 & 0.9662 \\
MCC      & 0.8523 & 0.8517 & 0.8561 \\
PR-AUC   & 0.9885 & 0.9892 & 0.9910 \\
ECE      & 0.0313 & 0.0934 & 0.0273 \\
Fake F1  & 0.8746 & 0.8747 & 0.8766 \\
Real F1  & 0.9765 & 0.9762 & 0.9772 \\
\bottomrule
\end{tabular}
\caption{BanglaBERT classification performance across three random seeds.}
\label{tab:appendix_banglabert_classification}
\end{table}

\begin{table}[H]
\centering
\setlength{\tabcolsep}{10pt}
\renewcommand{\arraystretch}{1.15}
\begin{tabular}{lccc}
\toprule
\textbf{Metric} & \textbf{Seed 22} & \textbf{Seed 33} & \textbf{Seed 42} \\
\midrule
P50 Latency (ms) & 37.35 & 36.81 & 37.07 \\
P95 Latency (ms) & 42.60 & 37.07 & 37.42 \\
Throughput (samples/s) & 1649.46 & 1699.17 & 1676.46 \\
Inference Peak VRAM (MB) & 1011.9 & 1011.9 & 1011.9 \\
Training Peak VRAM (MB) & 5141.6 & 5141.6 & 5141.6 \\

\bottomrule
\end{tabular}
\caption{BanglaBERT efficiency metrics across three random seeds.}
\label{tab:appendix_banglabert_efficiency}
\end{table}

\subsection{BanglaMamba Results}

\begin{table}[H]
\centering
\setlength{\tabcolsep}{10pt}
\renewcommand{\arraystretch}{1.15}

\begin{tabular}{lccc}
\toprule
\textbf{Metric} & \textbf{Seed 22} & \textbf{Seed 33} & \textbf{Seed 42} \\
\midrule
Macro F1 & 0.9036 & 0.9060 & 0.8992 \\
AUC-ROC  & 0.9539 & 0.9352 & 0.9369 \\
MCC      & 0.8075 & 0.8176 & 0.8007 \\
PR-AUC   & 0.9873 & 0.9776 & 0.9810 \\
ECE      & 0.0313 & 0.0479 & 0.0578 \\
Fake F1  & 0.8384 & 0.8403 & 0.8297 \\
Real F1  & 0.9688 & 0.9717 & 0.9687 \\
\bottomrule
\end{tabular}
\caption{BanglaMamba classification performance across three random seeds.}
\label{tab:appendix_banglamamba_classification}
\end{table}

\begin{table}[H]
\centering
\setlength{\tabcolsep}{10pt}
\renewcommand{\arraystretch}{1.15}
\begin{tabular}{lccc}
\toprule
\textbf{Metric} & \textbf{Seed 22} & \textbf{Seed 33} & \textbf{Seed 42} \\
\midrule
P50 Latency (ms) & 16.07 & 16.20 & 12.26 \\
P95 Latency (ms) & 17.87 & 20.35 & 14.06 \\
Throughput (samples/s) & 3744.20 & 3631.85 & 3493.65 \\
Inference Peak VRAM (MB) & 514.5 & 514.5 & 514.5 \\
Training Peak VRAM (MB) & 1145.4 & 1145.4 & 1145.4 \\
\bottomrule
\end{tabular}
\caption{BanglaMamba efficiency metrics across three random seeds.}
\label{tab:appendix_banglamamba_efficiency}
\end{table}

\subsection{CustomBERT Results}

\begin{table}[H]
\centering
\setlength{\tabcolsep}{10pt}
\renewcommand{\arraystretch}{1.15}
\begin{tabular}{lccc}
\toprule
\textbf{Metric} & \textbf{Seed 22} & \textbf{Seed 33} & \textbf{Seed 42} \\
\midrule
Macro F1 & 0.9051 & 0.9137 & 0.8982 \\
AUC-ROC & 0.9504 & 0.9579 & 0.9139 \\
MCC & 0.8108 & 0.8294 & 0.8010 \\
PR-AUC & 0.9845 & 0.9884 & 0.9700 \\
ECE & 0.0406 & 0.0467 & 0.0737 \\
Fake F1 & 0.8405 & 0.8545 & 0.8273 \\
Real F1 & 0.9696 & 0.9730 & 0.9691 \\
\bottomrule
\end{tabular}
\caption{CustomBERT classification performance across three random seeds.}
\label{tab:appendix_custombert_classification}
\end{table}

\begin{table}[H]
\centering
\setlength{\tabcolsep}{10pt}
\renewcommand{\arraystretch}{1.15}
\begin{tabular}{lccc}
\toprule
\textbf{Metric} & \textbf{Seed 22} & \textbf{Seed 33} & \textbf{Seed 42} \\
\midrule
P50 Latency (ms) & 36.90 & 36.84 & 37.36 \\
P95 Latency (ms) & 37.37 & 37.13 & 42.65 \\
Throughput (samples/s) & 1690.40 & 1696.59 & 1653.40 \\
Inference Peak VRAM (MB) & 1011.9 & 1011.9 & 1011.9 \\
Training Peak VRAM (MB) & 5141.6 & 5141.6 & 5141.6 \\
\bottomrule
\end{tabular}
\caption{CustomBERT efficiency metrics across three random seeds.}
\label{tab:appendix_custombert_efficiency}
\end{table}

\section{Data Integrity and Leakage Prevention}
\label{app:data_integrity}

To ensure that the reported results are not affected by data leakage, we performed a systematic check of the data preprocessing, partitioning, and evaluation pipeline. All corpus-level cleaning and deduplication operations were completed before dataset partitioning. The cleaned corpus was then divided using a stratified 80/10/10 train--validation--test split with a fixed random seed.

\subsection{Deduplication and Data Preprocessing}
\label{app:deduplication}

Each article was represented by the concatenation of its headline and body. Before deduplication, the text was normalized by removing HTML artifacts and URLs, applying Unicode NFC normalization, and collapsing redundant whitespace. Exact duplicates were then removed based on the complete normalized headline--body representation, retaining only the first occurrence. No approximate or semantic similarity matching was used.

After deduplication, articles containing fewer than 20 or more than 2,000 words were removed. Only after these preprocessing steps was the dataset partitioned into training, validation, and test sets. This ordering prevents identical normalized articles from occurring across different partitions.

\subsection{Leakage Prevention During Model Training and Evaluation}
\label{app:leakage_prevention}

The BanglaBERT tokenizer was used in its pretrained form and was not fitted or adapted to the study dataset. Class weights were computed exclusively from the training partition. During training, model checkpoints were selected solely according to validation Macro-F1, while the test set remained completely excluded from model selection and parameter updates. The final reported test results were obtained from the selected checkpoint in a separate evaluation pass.

These procedures provide safeguards against duplicate contamination, tokenizer fitting leakage, label-statistic leakage, and test-set-driven model selection.

\end{document}